\documentclass[letterpaper, 10 pt, conference]{ieeeconf}  % Comment this line out if you need a4paper

\IEEEoverridecommandlockouts                              % This command is only needed if 
\usepackage{bm,amsmath,amsfonts,amssymb,dsfont} % math
\usepackage{multirow} % for table
\usepackage{graphicx} % for figures
\graphicspath{{images/}}

\usepackage{hyperref}
\hypersetup{
    colorlinks=true,
    linkcolor=blue,
    filecolor=magenta,      
    urlcolor=blue,
}

\newcommand{\NewR}{\ensuremath{\mathds{R}}}
\newcommand{\mat}[1]{{\ensuremath{{\mathbf{#1}}}}}

\usepackage{siunitx} % degree

\DeclareMathOperator*{\argmax}{argmax} % thin space, limits underneath in displays
\title{\LARGE \bf
Transferring Experience from Simulation to the Real World for Precise Pick-And-Place Tasks in Highly Cluttered Scenes
}

\author{

\begin{small}
Kilian Kleeberger$^{1}$,
Markus Völk$^{1}$,
Marius Moosmann$^{1}$,
Erik Thiessenhusen$^{1}$,
Florian Roth$^{1}$,
Richard Bormann$^{1}$,
Marco F. Huber$^{2,3}$
\end{small}

\thanks{$^{1}$Department Robot and Assistive Systems, Fraunhofer Institute for Manufacturing Engineering and Automation IPA,
        Nobelstra{\ss}e~12, 70569 Stuttgart, Germany
        {\tt\small kilian.kleeberger@ipa.fraunhofer.de}}%
\thanks{$^{2}$Center for Cyber Cognitive Intelligence (CCI), Fraunhofer Institute for Manufacturing Engineering and Automation IPA,
        Nobelstra{\ss}e~12, 70569 Stuttgart, Germany
        {\tt\small marco.huber@ipa.fraunhofer.de}}%
\thanks{$^{3}$Institute of Industrial Manufacturing and Management IFF, University of Stuttgart,
        Allmandring~35, 70569 Stuttgart, Germany
        {\tt\small marco.huber@ieee.org}}%
}

\begin{document}

\maketitle
\thispagestyle{empty}
\pagestyle{empty}

%%%%%%%%%%%%%%%%%%%%%%%%%%%%%%%%%%%%%%%%%%%%%%%%%%%%%%%%%%%%%%%%%%%%%%%%%%%%%%%%
%%%%%%%%%%%%%%%%%%%%%%%%%%%%%%%%%%%%%%%%%%%%%%%%%%%%%%%%%%%%%%%%%%%%%%%%%%%%%%%%
%%%%%%%%%%%%%%%%%%%%%%%%%%%%%%%%%%%%%%%%%%%%%%%%%%%%%%%%%%%%%%%%%%%%%%%%%%%%%%%%
\begin{abstract}
In this paper, we introduce a novel learning-based approach for grasping known rigid objects in highly cluttered scenes and precisely placing them based on depth images. Our Placement Quality Network (PQ-Net) estimates the object pose and the quality for each automatically generated grasp pose for multiple objects simultaneously at 92~fps in a single forward pass of a neural network. All grasping and placement trials are executed in a physics simulation and the gained experience is transferred to the real world using domain randomization. We demonstrate that our policy successfully transfers to the real world. PQ-Net outperforms other model-free approaches in terms of grasping success rate and automatically scales to new objects of arbitrary symmetry without any human intervention.
\end{abstract}

% ((very/highly)) precise object placement based on depth images, know object model
% precise placement of objects.
%to grasp pose success prediction

%that estimates all required quantities simultaneously.
%predict the "quality of a discrete set of grasp candidates simultaneously"
%We simultaneously estimate the pose of multiple objects in the image together with the quality for each grasp pose defined on the objects and select the highest ranked grasp.

% knowledge/

%We demonstrate that our approach can be deployed to the real world
%Despite being entirely trained on simulated data,

%works on real-world data
% compare/comparison with other approaches

%/involvement.
% configure for new 3D object models (for precise placement).
% can successfully be transferred to applications.

%%%%%%%%%%%%%%%%%%%%%%%%%%%%%%%%%%%%%%%%%%%%%%%%%%%%%%%%%%%%%%%%%%%%%%%%%%%%%%%%
%%%%%%%%%%%%%%%%%%%%%%%%%%%%%%%%%%%%%%%%%%%%%%%%%%%%%%%%%%%%%%%%%%%%%%%%%%%%%%%%
%%%%%%%%%%%%%%%%%%%%%%%%%%%%%%%%%%%%%%%%%%%%%%%%%%%%%%%%%%%%%%%%%%%%%%%%%%%%%%%%
\section{Introduction}

% "Why was the study done?"
% "Give the reader a sufficient background"

% Motivation...
% Robotic grasping and manipulation is a longstanding challenge... [QT-Opt]
For robots to work safely and effectively, they must be aware of their environment. One aspect of this is the estimation of the pose of the objects in the scene to be able to avoid collisions and allow robust grasping and manipulation of the components.
% Scenes of Many Parts in Bulk
6D object pose estimation (OPE) and grasp planning in highly cluttered scenes based on a single depth image is
%hard/
challenging because of sensor noise, incomplete information, and uncertainties about the state of the environment. Furthermore, the robot has to reason
%because it require reasoning
on how to manipulate the objects because selecting the wrong object and grasp pose can result in failed grasps.

%%%%%%%%%%%%%%%%%%%%%%%%
% In-hand localization %
%%%%%%%%%%%%%%%%%%%%%%%%

% Model-free approaches for defined placement of rigid objects: in-hand localization can be done afterwards. still, the placement might not be kinetically feasible; with our approach this is checked beforehand (should be checked beforehand)

% model-free grasping and in-hand localization --> object might not be placeable due to kinematic constraints
% --> consider early on during the grasp planning process

% Contrary to the research trend towards model-free robotic grasping, we focus on how to incorporate model knowledge into autonomous robot system.
% (Because a CAD model is given in many/most industrial scenarios)
% In this work, we incorporate model knowledge into learning-based grasping/picking systems. %(model is given in many industrial scenarios)
%
%
%
% allows to place the grasped object in a defined pose (what model-free approaches cannot do)

% approaches to robotic grasping either directly predict a grasp pose given sensor data and other do a prior pose estimation step before
% planning a suitable path for grasping

% very little works are available tackling model-based approaches...

\begin{figure}[tb]
\centering
\includegraphics[scale=0.25]{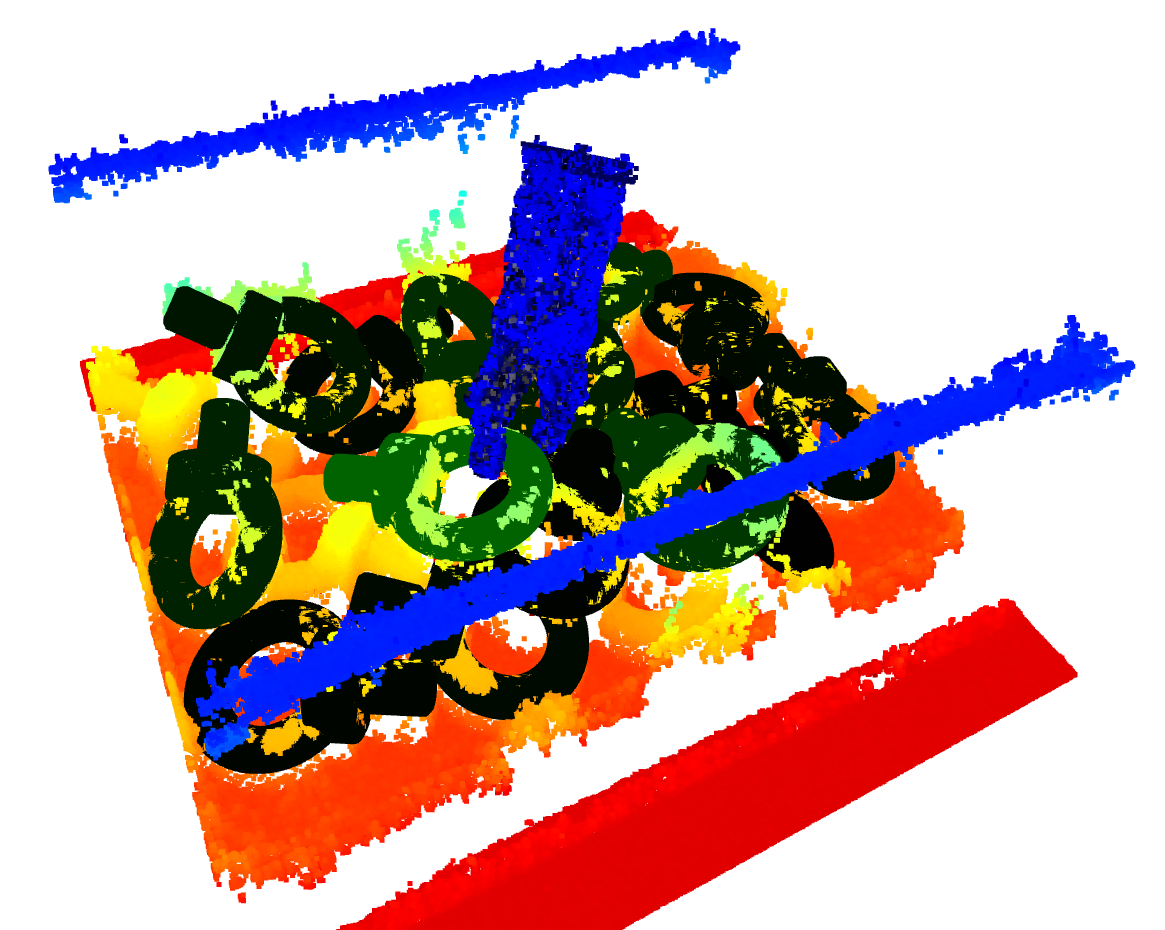}
\caption{Estimated object poses of our approach on real-world data after ICP refinement. The greener the object, the more certain the model is that the object can be grasped safely. The gripper (blue) indicates the top ranked grasp based on our
%grasping
policy.
%... indicates suitable grasp poses. (exemplary visualize some top ranked grasps)
}
\label{fig:result_real_world_data}
\end{figure}

%%%%%%%%%%%%%%%%%%%%%%%%
%%% Cluttered scenes %%%
%%%%%%%%%%%%%%%%%%%%%%%%

% Other learning-based approaches only consider scenarios of already separated objects [DOPE].

\begin{figure}[tb]
\centering
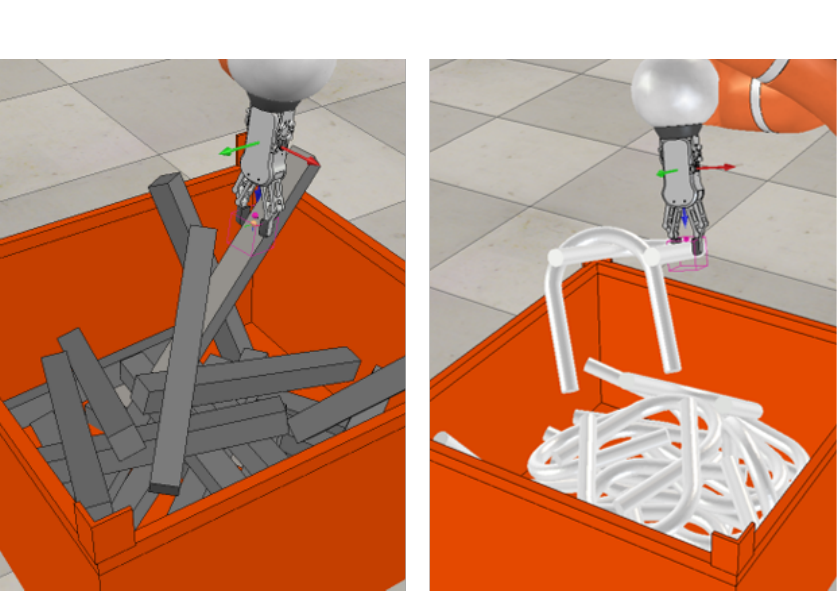
\caption{Failure cases
%/ Challenges
for picking tasks in cluttered scenes:
(left) The robot picks an object (IPABar) which moves relative to the gripper during
%grasping/
lifting due to overlaps with other objects. The object cannot be placed precisely anymore. The objects in the bin have to be picked in a defined order.
(right) The robot picks an object (IPAUBolt) which is entangled with another object.
%entanglement; entangled objects (IPAUBolt);
Therefore, a goal is to select objects
%for grasping
which do
%not result in a high amount of unrest in the bin and
do not entangle
% or jam (?; verklemmen?) (avoid hooking or jamming)
with other objects.}
\label{fig:Motivation}
\end{figure}

%DOPE~\cite{DOPE}
Works such as~\cite{DOPE,IanLenz.2015,JosephRedmon.2015,Jacquard} focus on robotic grasping and manipulation tasks in scenarios with limited clutter
% simple scenarios (with limited clutterd; no dense clutter)
which do not require a defined picking order of the objects.
%
% usually simple scenarios (singulated objects); no dense clutter
%
%analytical approaches only check the collision-free reachability of grasp poses
%do not consider the actual grasping of the object in simulation
%which is important for cluttered scenes and object of complex shape.
%
% does not answer:
%Which object is best to pick next? which grasp pose should be used? Does the proposed grasp result in a highly precise placement (reliability)?
%--> all of this is not answered by analytical path planning methods
%
Simply selecting collision-free and kinematically feasible grasps in highly cluttered scenes~\cite{FPS_GPC_1,FPS_GPC_2}, might lead to a movement of the object relative to the gripper which prevents a precise placement without additional in-hand localization and entanglements with other objects for complex object geometries as visualized in Fig.~\ref{fig:Motivation}.
% unwisely picked grasps:
% object might move relative to the gripper and no precise placement is possible 
% object that is picked high potential to entangle with other objects
% In highly cluttered scenes, grasps for precise placement often fail due to a movement of the object relative to the gripper objects, entanglements, or jams
% Verklemmungen
% with other objects (which are hard to determine analytically)
In this paper, we tackle these challenges by providing a novel learning-based approach for grasp pose evaluation in
%highly cluttered scenes
scenes of many parts in bulk.

% Objects not suitable for grasping might be detected with a very high score --> common approaches start searching for grasp poses at that objects
% --> We: Unify everything in one framework

%%%%%%%%%%%%%%%%%%%%%%%%%%%%%%%
%%% Automatic configuration %%%
%%%%%%%%%%%%%%%%%%%%%%%%%%%%%%%

% Model-free approaches to robotic grasping aim for generalization to novel/unseen objects (grasping). We provide a model-based approach, which automatically configures to new objects for precise object placement, i.e. the provided solution does not require any manual object-specific tuning.

%%%%%%%%%%%%%%%%%%%%%%%%%%%%%%
%%% Use physics simulation %%%
%%%%%%%%%%%%%%%%%%%%%%%%%%%%%%
% Other works
Approaches to robotic grasping and manipulation usually rely on datasets consisting of human-labeled grasps~\cite{IanLenz.2015,JosephRedmon.2015,GG-CNN}, which are tedious to get, or on physical grasp outcomes where data collection can take several months~\cite{Google_2016_Learning_Hand-Eye_Coordination,Google_2018_Learning_Hand-Eye_Coordination,Supersizing_Self_Supervision,QT_Opt}.
%Our approach: Use simulation and Sim-to-Real Transfer
In this work, we execute each predefined grasp pose in a physics simulation and
% Profit/Profits:
transfer the gained
%(knowledge/)
experience
%from simulation
to the real world using domain randomization~\cite{Domain_Randomization} to increase generalization.
%%%to make it generalize in new unseen situations.
% Our approach works on real-world data without requiring any (additional) fine tuning (Zero-Shot Transfer [XX]).
Our approach directly transfers from simulation to the real world (see Fig.~\ref{fig:result_real_world_data}).
% avoid configuration effort and manual tuning...
% the system hast to
% configures itself for precise object placement without any human intervention.
% (ohne jegliches menschliches Zutun)
% In this work, we execute the grasps in a physics simulation and try to transfer the knowledge form simulation to the real world and make it (with the aim to make it) generalize in new unseen situations.

Learning-based approaches to robotic grasping~\cite{Dex-Net_1.0,Dex-Net_2.0,Dex-Net_3.0,Google_2016_Learning_Hand-Eye_Coordination,Google_2018_Learning_Hand-Eye_Coordination,QT_Opt} usually rely on top-down grasps and cannot be used for bin-picking due to collisions when attempting to grasp objects close to the border of the bin. Analytical approaches for \mbox{pick-and-place} tasks in cluttered scenes require an object-specific configuration and tuning until a satisfactory
%/satisfiable
system performance is reached, which limits the scalability~\cite{FPS_GPC_1,Config_Effort_Bin_Picking,ReviewArticleSpringer_KBK,paper_TEM,IPA_Diss_ThomasLedermann.2012,IPA_Diss_MatthiasPalzkill.2014}. \mbox{PQ-Net} configures automatically based on a given object model and does not require any human intervention.

%%%%%%%%%%%%%%%%%%%%%%%%%%%%%%
%%% Extension in this work %%%
%%%%%%%%%%%%%%%%%%%%%%%%%%%%%%
% is is an (model-based) object pose estimation approach for robotic grasping --> why not extend it to robotic grasp evaluation / quality assessment?

%Motivated/
%Inspired by the success of single shot approaches (for 6D object pose estimation), we (go further and) extend a state-of-the-art approach for OPE
%approach / framework
%to grasp success prediction.
% Success: 
% - ...

%In this paper,
Inspired by the success of single shot approaches~\cite{YOLOv1,YOLOv2,SSD,Tekinetal.},
%(for 6D object pose estimation)
we go further and extend OP-Net~\cite{OP-Net}, a single shot approach for OPE outperforming the winner of the ``\textit{Object Pose Estimation Challenge for Bin-Picking}'' at IROS 2019\footnote{\url{http://www.bin-picking.ai/en/competition.html}} on the Sil\'{e}ane dataset~\cite{Bregier_SymmetryAwareEvaluation}. To the best of our knowledge, we are the first extending a single shot approach for OPE to
%grasp planning
grasp success prediction in a joint framework.

%%%%%%%%%%%%%%%%%%%%%%%%%%%%%%%%%%%
%%% \section{Problem Statement} %%%
%%%%%%%%%%%%%%%%%%%%%%%%%%%%%%%%%%%
Based on a single depth image, \mbox{PQ-Net} predicts the object pose $P \in \mathrm{SE}(3)$ relative to the camera coordinate system and outputs a success estimate for a set of predefined grasps $\mathcal{G}$ defined relative to the object coordinate system.
% (body-fixed coordinate system)
%
% specifying whether selecting/executing $G$ results in a successful placement or not
%
% predict the success label for each grasp pose $G \in \mathrm{SE}(3)$ element G (500 grasps)
%(((which result in)))
% for a precise placement of the object 
%
Furthermore, we introduce graspability metrics
%(quality metrics)
which allow a
%material-friendly /
gentle component removal
%(materialschonende Bauteilentnahme)
and avoid entanglements.

% select a successful grasp while avoiding entanglements and minimizing the / avoiding unrest caused in the scene by picking the object (avoid movement of the object relative to the gripper and a

% 3D object model $\mathcal{M}$

%%%%%%%%%%%%%%%%%%%%%
%%% Contributions %%%
%%%%%%%%%%%%%%%%%%%%%
In summary, the main contributions of this work are:
% In summary, our work has the following key contributions:
\begin{itemize}
\item Extension of a single shot approach
%(state-of-the-art approach)
for object pose estimation to grasp pose success prediction suitable for precise object placement
% provide a learning-based approach for precise object placement
\item Novel metrics for the assessment of the graspability of objects in highly cluttered scenes
% Extend the “Fraunhofer Bin-Picking dataset” with success labels for each grasp pose (success: collision-free reachability)
%\item Scalable framework that allows to importing an object model of arbitrary symmetry to a physics simulation only to letting a robot learn to precisely place objects autonomously %(provide a framework for the
\item Scalable system, which enables a robot to learn how to place objects precisely on the basis of an object model of arbitrary symmetry (automatic configuration)
% of precise \mbox{pick-and-place} tasks
%\item Properly consider object symmetries during data generation (grasping and placing) and in the loss function (regression of the angles)
\item Extension of the Fraunhofer IPA~\cite{FraunhoferIPABinPickingDataset} and Sil\'{e}ane~\cite{Bregier_SymmetryAwareEvaluation} datasets with grasp annotations
and provide data for two new objects. All datasets are publicly available at
%\url{https://owncloud.fraunhofer.de/index.php/s/efdHnOL2b26OcqG}. % link data of submission for review
%\url{https://owncloud.fraunhofer.de/index.php/s/975EH0WxWYG0noY}.
\url{http://www.bin-picking.ai/en/dataset.html}.
%(do separate paper?!); Extend the "Fraunhofer IPA Bin-Picking dataset" with two novel objects (IPABar and IPAUBolt) (separate paper?!)
\end{itemize}

The paper is structured as follows. The next section reviews related work. In Section~\ref{sec:approach} the proposed approach is described. Experimental evaluations are provided in Section~\ref{sec:evaluation}. Pros and cons of our approach are discussed in Section~\ref{sec:discussion}. The paper closes with a conclusion.

% Research in Bin-Picking [IROS 2019, ...]
% All datasets are available in the format of the "Fraunhofer IPA Bin-Picking dataset" at ...
% Google2016 real world --> limited scalability; therefore: Train in simulation and use ... to transfer the model to the real world.
% analytisch schwer bestimmbar, ob Griff erfolgreich wird oder nicht...

%doing rich set of pre-grasp manipulations for robust grasping [QT-Opt], tossing of object [tossingbot], ... (focusing on generalization to novel objects) but precise placement hast been ignored...

%%%%%%%%%%%%%%%%%%%%%%%%%%%%%%%%%%%%%%%%%%%%%%%%%%%%%%%%%%%%%%%%%%%%%%%%%%%%%%%%
%%%%%%%%%%%%%%%%%%%%%%%%%%%%%%%%%%%%%%%%%%%%%%%%%%%%%%%%%%%%%%%%%%%%%%%%%%%%%%%%
%%%%%%%%%%%%%%%%%%%%%%%%%%%%%%%%%%%%%%%%%%%%%%%%%%%%%%%%%%%%%%%%%%%%%%%%%%%%%%%%
\section{Related Work}
Methods for robotic grasping can roughly be categorized in analytical
%/analytic
and
%empirical (or
data-driven methods~\cite{Sahbani_2012,Bohg_2014,ReviewArticleSpringer_KBK}.

%%%%%%%%%%%%%%%%%%%%%%%%%%%%%%%%%%%%%%%%%%%%%%%%%%%%%%%%%%%%%%%%%%%%%%%%%%%%%%%%
%%%%%%%%%%%%%%%%%%%%%%%%%%%%%%%%%%%%%%%%%%%%%%%%%%%%%%%%%%%%%%%%%%%%%%%%%%%%%%%%
%%%%%%%%%%%%%%%%%%%%%%%%%%%%%%%%%%%%%%%%%%%%%%%%%%%%%%%%%%%%%%%%%%%%%%%%%%%%%%%%
\subsection{Analytical Approaches} % Classical % to robotic grasping

Analytical approaches (model-based approaches) use an object model with predefined grasps. First off, they localize the object in the scene~\cite{BinPicking_6D_OPE_IROS_2019}. Based on this, they try to find a collision-free and kinematically feasible path
%towards the object
for grasping and placing~\cite{FPS_GPC_1,FPS_GPC_2}.
% ((using heuristics))
% ((using heuristic search~\cite{FPS_GPC_2}))
% bp3: GPC: möglichst schnell geeigneten Kandidaten finden und dann Griff ausführen --> zwar kollisionsfrei zugänglich und kinematically fasible aber oft nicht robuster Griff
%
% These approaches do not scale to novel objects.
%
% bp3 requires:
% - Manual tuning of grasp poses
% - Manual tuning of prioritization for each grasp pose
% - Manual definition of clear zones
%
Especially for highly cluttered scenes, they require significant effort for manually tuning suitable grasp poses and grasp priorities to reach a satisfactory
%satisfyable
system performance,
%a huge amount of manual tuning for the definition of suitable grasp poses together with a priority (usually done manually) until the system performance is satisfied,
limiting the scalability to new
%/novel
objects~\cite{FPS_GPC_1,Config_Effort_Bin_Picking,ReviewArticleSpringer_KBK,paper_TEM,IPA_Diss_ThomasLedermann.2012,IPA_Diss_MatthiasPalzkill.2014}. Usually, the grasp poses are prioritized independent of the object pose. Furthermore, zones on the object can be specified where no measurement point of the 3D point cloud should be contained in order to pick the candidate next. This can be used to specify a
% defined
picking order of the localized objects in the scene.

%Classical/
%Analytical approaches [XX, XX] require
%a high amount of manual tuning
%expert knowledge for the OPE with template [XX] or feature [XX] based matching and a high amount of manual tuning in terms of grasp pose definition and prioritization of the grasp poses for a robust localization and picking of the objects.

% "known 3D objects labeled with grasps and quality metrics" (prioritization) % (Dex-Net)
% "executing the highest quality grasp" % (Dex-Net)

%%%%%%%%%%%%%%%%%%%%%%%%%%%%%%%%%%%%%%%%%%%%%%%%%%%%%%%%%%%%%%%%%%%%%%%%%%%%%%%%
%%%%%%%%%%%%%%%%%%%%%%%%%%%%%%%%%%%%%%%%%%%%%%%%%%%%%%%%%%%%%%%%%%%%%%%%%%%%%%%%
%%%%%%%%%%%%%%%%%%%%%%%%%%%%%%%%%%%%%%%%%%%%%%%%%%%%%%%%%%%%%%%%%%%%%%%%%%%%%%%%
\subsection{Learning-based Approaches}
%\subsection{Model-free Approaches}
%\subsection{Model-free Robotic Grasping}

%%%%%%%%%%%%%%%%%%%%%%%%%%%%%%%
%%% Robotic Grasp Detection %%%
%%%%%%%%%%%%%%%%%%%%%%%%%%%%%%%

Approaches to Robotic Grasp Detection estimate oriented rectangles in the input image which represent a grasp configuration for parallel jaw grippers~\cite{rectangle_representation_ICRA_2011}.
Public datasets are the Cornell Grasping Dataset~\cite{IanLenz.2015} providing 1,035 manually annotated samples of 280 objects and the Jacquard Dataset~\cite{Jacquard}
% utilizes 11k objects and a physics simulation to generate a large-scales synthetic dataset,
with over 50,000 synthetic samples on a large diversity of objects (11,000), each with multiple labeled grasps.
% images labeled with grasps parametrized by oriented bounding boxes
% Large-scale synthetic dataset with ground truth
% Contains a large diversity of objects (11k), each with multiple labeled grasps.

%Redmon et al.
MultiGrasp~\cite{JosephRedmon.2015} uses the Cornell dataset to train a neural network to predict oriented rectangles (bounding boxes) in an image together with a confidence and
% SingleGrasp Detection~\cite{JosephRedmon.2015}
% MultiGrasp Detection~\cite{JosephRedmon.2015}
makes local predictions based on global information by discretizing the output in $S \times S$ grid cells.
%and predict an oriented bounding box together with a confidence measure for each grid cell.
This work led to the
%(famous)
YOLO~\cite{YOLOv1,YOLOv2} approach for object detection.
With a two-stage system that first samples grasp candidates and ranks them using neural networks, Lenz et al.~\cite{IanLenz.2015} demonstrated that this parameterization (oriented rectangles) can be used for real-world robotic grasping tasks.

%%%%%%%%%%%%%%
%%% GG-CNN %%%
%%%%%%%%%%%%%%
GG-CNN~\cite{GG-CNN, GG-CNN_2} predicts a quality and configuration of grasps at every
%each?!
pixel of the input image using a lightweight convolutional neural network trained on the Cornell and Jacquard dataset~\cite{Jacquard}.
%GG-CNN "predicts the quality and pose of grasps at every pixel."
The generated antipodal grasps that are executed closed-loop and allow grasping in cluttered scenes and non-static environments.
% (using top-down grasps)

%%%%%%%%%%%%
%%% ToDo %%%
%%%%%%%%%%%%
%predict the "quality of a discrete set of grasp candidates simultaneously"~\cite{Johns.2016}
%[Johns et al. IROS 2016]
% also done by "superising self-supervision according to GG-CNN --> check that

%%%%%%%%%%%%%%%
%%% Dex-Net %%%
%%%%%%%%%%%%%%%
Dex-Net makes use of large scale synthetic data collection for learning grasping policies for parallel jaw~\cite{Dex-Net_1.0, Dex-Net_2.0} and suction grippers~\cite{Dex-Net_3.0}
% or both
using analytic metrics. The sampled grasps are ranked using a neural network which gets a cropped depth image and grasp candidate as input.
Dex-Net observes a local image patch, and is not designed to execute grasps in a defined order or avoid entangled objects due to missing global scene information.
% by sampling grasp poses and using.
% a combinations [xx], and a fully convolution version [xx].

%%%%%%%%%%%%%%%%%%%%%%%%%%%%%%%%%%%%%%%%%
%%% Hand-Eye Coordination Google 2016 %%%
%%%%%%%%%%%%%%%%%%%%%%%%%%%%%%%%%%%%%%%%%
% Learning Hand-Eye Coordination
Levine et al.~\cite{Google_2016_Learning_Hand-Eye_Coordination,Google_2018_Learning_Hand-Eye_Coordination} parallelizes the real-word data collection to up to 14 robot and collect 800,000 samples in two months for robust grasping.
%%%%%%%%%%%%%%
%%% QT-Opt %%%
%%%%%%%%%%%%%%
QT-Opt~\cite{QT_Opt} makes use of reinforcement learning
%[XX]
to train robotic grasping and manipulation policies based on self-supervision on real-world systems.
%((for picking objects))
Because of the time-consuming and hardware demanding data collection procedure, works such as GraspGAN~\cite{KonstantinosBousmalis.2018} or RCAN~\cite{RCAN} focus on reducing the need of real-world data collection.

%%%%%%%%%%%%%%%%%%%
%%% Limitations %%%
%%%%%%%%%%%%%%%%%%%
% all above motioned works:
While all these aforementioned model-free approaches to robotic grasping show promising results, they do not provide a solution for a precise placement of the objects and only consider pick-and-drop tasks using top-down grasps (4D). Using grasps in this grasp representation
%(top-down grasps)
has limitations, e.g., for bin-picking due to collisions with the bin when attempting to grasp objects at the border. Therefore, works such as
%Furthermore
\cite{Google_2016_Learning_Hand-Eye_Coordination,Google_2018_Learning_Hand-Eye_Coordination,Dex-Net_1.0,Dex-Net_2.0,QT_Opt,Morrison_MultiViewPicking,Berscheid.2019} use bins with slanted or no high
%/sloped
bin walls to ensure that the top-down grasps
%(still/can)
work.
Furthermore, it is not possible to blindly move into the bin for data collection due to damaging the gripper because of unknown fill levels of the bins. Picking multiple objects is
often
also considered as a successful grasp~\cite{Google_2016_Learning_Hand-Eye_Coordination,Google_2018_Learning_Hand-Eye_Coordination,QT_Opt}.

% Disadvantage:
% - Might grasp multiple objects (usually considered as a successful grasp) / While grasping multiple objects is typically considered as a successful grasp [Levine][QT-Opt][...] we also log Verhakungen...

% Model-free approaches:
% - Object classification
% - Precise placement for rigid objects
% - Placement might not be kinematically feasible (consider this early / during grasp pose selection)

%%% Other:
% Failure of approaches such as GG-CNN or Dex-Net where a precise picking order is required...
% due to the local evaluation of Dex-Net and the training of GG-CNN on single isolated objects (Cornell Grasping dataset, Jacquard dataset) and no objects of many scenes in bulk

%%%%%%%%%%%%%%%%%%%%%%%%%%%%%%%%%%%%%%%%%%%%%%%%%%%%%%%%%%%%%%%%%%%%%%%%%%%%%%%%
%%%%%%%%%%%%%%%%%%%%%%%%%%%%%%%%%%%%%%%%%%%%%%%%%%%%%%%%%%%%%%%%%%%%%%%%%%%%%%%%
%%%%%%%%%%%%%%%%%%%%%%%%%%%%%%%%%%%%%%%%%%%%%%%%%%%%%%%%%%%%%%%%%%%%%%%%%%%%%%%%
\section{Placement Quality Network} % Approach
% GQ-Net % Grasp Quality Network % deprecated
% PQ-Net % Placement Quality Network
\label{sec:approach}
% This part should not be bin-picking specific! (General approach for 6D OPE)
In this section, we describe the routine for automatically generating grasp poses for object models, the process of data generation for training our neural network using a physics simulation, the proposed definitions for the graspability of objects, the parameterization of the network's output, the loss function, the network architecture, and the training procedure together with the technique for a robust transfer of the model from simulation to the real world. Fig.~\ref{fig:Overview} shows an overview of our approach.

%%%%%%%%%%%%%%%%%%%%%%%%%%%%%%%%%%%%%%%%%%%%%%%%%%%%%%%%%%%%%%%%%%%%%%%%%%%%%%%%
%%%%%%%%%%%%%%%%%%%%%%%%%%%%%%%%%%%%%%%%%%%%%%%%%%%%%%%%%%%%%%%%%%%%%%%%%%%%%%%%
%%%%%%%%%%%%%%%%%%%%%%%%%%%%%%%%%%%%%%%%%%%%%%%%%%%%%%%%%%%%%%%%%%%%%%%%%%%%%%%%
\subsection{Automatic Grasp Pose Generation}

% plans robust grasps analytically using full knowledge of 3D object geometry

To avoid the need for manually defining grasp poses $G \in \mathrm{SE}(3)$ on the object model, we provide a method that
%/to
automatically generates a set of grasp poses $\mathcal{G}$ for common gripper types such as parallel jaw, suction, and magnetic grippers based on a given 3D object model.
Each grasp $G \in \mathcal{G}$ is represented by $(R; t) \in \mathrm{SE}(3)$ where $R \in \mathrm{SO}(3)$ and $t \in \NewR^3$ are the rotation and translation of grasp $G$.

%(CAD model)
As a first step of our technique, points are sampled on the surface of the object. For parallel jaw grippers, we check the distance between all pairs of points to verify whether it is smaller than the opening distance of the gripper,
%every pair proposes a grasp
%Secondly, we 
filter the candidates using the normal information of the 3D points, discretize the rotation around the straight line between any two points in
%$\frac{\ang{360}}{16}$
\ang{20} steps, and
% (further)
finally filter the candidates with a collision check using the gripper model. For suction and magnetic grippers, we sample grasp poses by evaluating the flatness of the object locally. Depending on the shape of the gripper, we define cylinder or cuboid elements, which should and should not contain points of the object model while also taking surface normals into account.

\begin{figure}[tb]
\centering
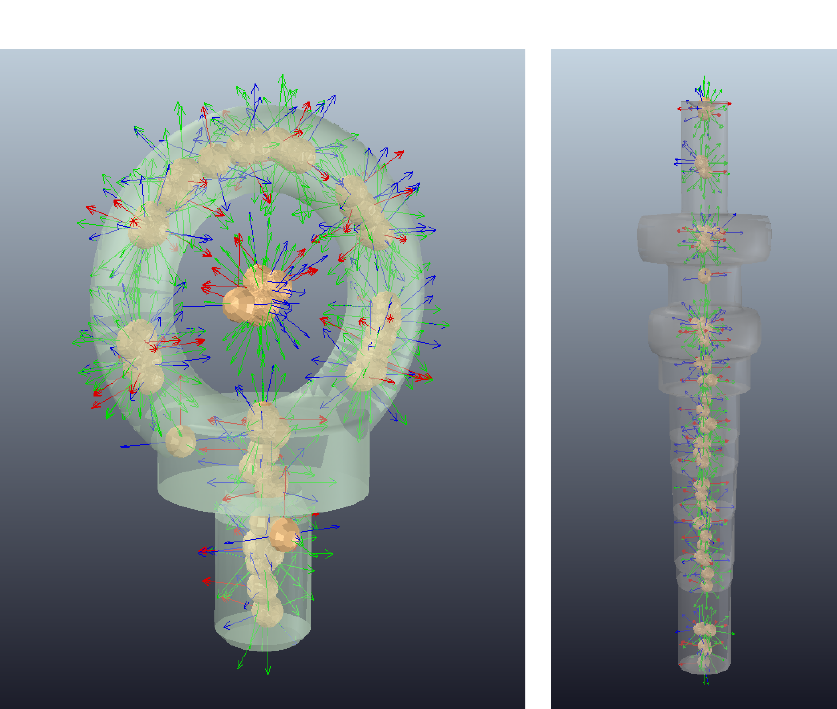
\caption{Automatically generated
%grasps /
grasp poses exemplary visualized for the IPARingScrew (left) and IPAGearShaft (right) for a parallel jaw gripper. Objects are taken from the Fraunhofer IPA dataset~\cite{FraunhoferIPABinPickingDataset}.}
\label{fig:GraspPoses}
\end{figure}

% Clustering
The proposed procedure results in a high number of grasp poses.
% resulting in a high number of grasp poses. 
We make use of unsupervised learning to reduce the amount of data while keeping a high diversity in terms of position and orientation. We apply partitioning around medoids (PAM)~\cite{PAM} clustering
% We then apply clustering (Unsupervised Learning); PAM
to reduce the number of grasps to approximately 500. Fig.~\ref{fig:GraspPoses} exemplary shows automatically generated grasp poses using our technique.

% which can be used for generating grasps of common gripper types such as parallel jaw, suction, and magnetic gripper. While for the latter two we simply search for "~ ebene Flächen" at the 3D object model.

%- Parallel jaw gripper
%- Suction gripper
%- Magnetic gripper

%\begin{figure*}[thpb]

%%%%%%%%%%%%%%%%%%%%%%%%%%%%%%%%%%%%%%%%%%%%%%%%%%%%%%%%%%%%%%%%%%%%%%%%%%%%%%%%
%%%%%%%%%%%%%%%%%%%%%%%%%%%%%%%%%%%%%%%%%%%%%%%%%%%%%%%%%%%%%%%%%%%%%%%%%%%%%%%%
%%%%%%%%%%%%%%%%%%%%%%%%%%%%%%%%%%%%%%%%%%%%%%%%%%%%%%%%%%%%%%%%%%%%%%%%%%%%%%%%
\subsection{Physics Simulation for Data Generation}

\begin{figure*}[thpb]
\begin{footnotesize}
\centering
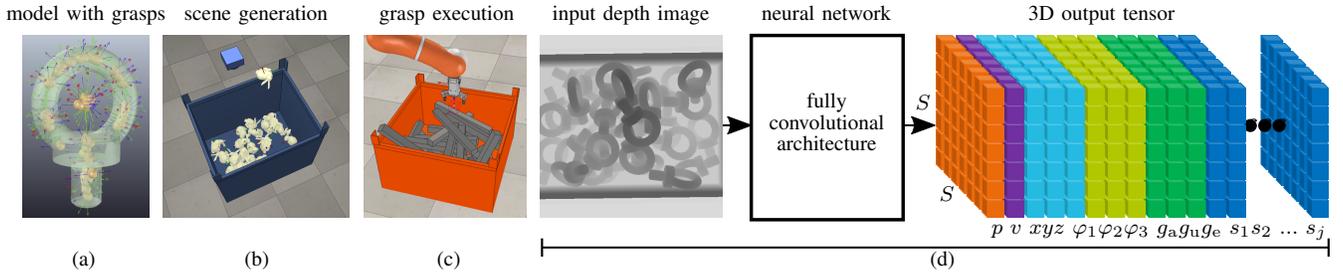
\caption{Overview of our approach: (a) 3D object model with automatically generated grasp poses (b) Physics simulation for scene generation (c) Physics simulation for grasp execution with a robot
% execute all (automatically generated / predefined) grasp poses in a physics simulation
(d) We train a deep neural network for 6D object pose estimation and grasp pose success estimation to transfer the knowledge gained in simulation to the real world using domain randomization~\cite{Domain_Randomization}. The output of our network is a 3D tensor comprising estimates of the probability $\hat{p}$, visibility $\hat{v}$, positions $\hat{x}$, $\hat{y}$, $\hat{z}$, Euler angles $\hat{\varphi}_1$, $\hat{\varphi}_2$, $\hat{\varphi}_3$, graspabilies $\hat{g}_\mathrm{a}$, $\hat{g}_\mathrm{u}$, $\hat{g}_\mathrm{e}$, and success $\hat{s}_j$ for each grasp pose $G_j \in \mathcal{G}$.}
\label{fig:Overview}
\end{footnotesize}
\end{figure*}

%%%%%%%%%%%%%%%%%%%%%%%%%%%%%%%%%%%%%%%%%%%%%%%%%%%%%%%%%%%%%%%%%%%%%%%%%%%%%%%%
%%%%%%%%%%%%%%%%%%%%%%%%%%%%%%%%%%%%%%%%%%%%%%%%%%%%%%%%%%%%%%%%%%%%%%%%%%%%%%%%
%%%%%%%%%%%%%%%%%%%%%%%%%%%%%%%%%%%%%%%%%%%%%%%%%%%%%%%%%%%%%%%%%%%%%%%%%%%%%%%%
\subsubsection{Scene Generation} % Fill Bins (and Data Logging)
We use the physics simulation \mbox{V-REP} / CoppeliaSim~\cite{V-REP} to create scenes with a high amount of clutter. % and ...
These scenarios are challenging because the robot has to avoid collisions with other objects in the scene and carefully select which object to pick next.
% Similar to / As done in /
Analogous to the Sil\'{e}ane~\cite{Bregier_SymmetryAwareEvaluation} and Fraunhofer IPA~\cite{FraunhoferIPABinPickingDataset} datasets, we drop objects in a random position and orientation above a bin to generate chaotic scenes typical for bin-picking.
We save the RGB image, depth image, and segmentation masks together with the visibility $v \in [0,1]$ and pose $P \in \mathrm{SE}(3)$ for each object in the scene. The number of objects that are dropped is increased iteratively until a predefined drop limit is reached, resulting in a uniform distribution over different fill levels of the bin (see also~\cite{FraunhoferIPABinPickingDataset}).

%%%%%%%%%%%%%%%%%%%%%%%%%%%%%%%%%%%%%%%%%%%%%%%%%%%%%%%%%%%%%%%%%%%%%%%%%%%%%%%%
%%%%%%%%%%%%%%%%%%%%%%%%%%%%%%%%%%%%%%%%%%%%%%%%%%%%%%%%%%%%%%%%%%%%%%%%%%%%%%%%
%%%%%%%%%%%%%%%%%%%%%%%%%%%%%%%%%%%%%%%%%%%%%%%%%%%%%%%%%%%%%%%%%%%%%%%%%%%%%%%%
\subsubsection{Grasping} % (and Data Logging)
Using the filled bins, we loop over each (automatically generated) grasp pose for all objects in the scene.
% (1st step)
First off, we check the collision of the gripper at every grasp pose with the environment (other objects and the bin).
% no collision: path planning
In case no collision occurs, we try to find a kinematically feasible robot configuration and plan a collision-free path to the grasp pose using the OMPL~\cite{OMPL} module integrated in \mbox{V-REP} / CoppeliaSim~\cite{V-REP}.

In case a suitable path was found, we execute the grasp and place the object at the defined target pose. We log whether an object is in the gripper after lifting and after placement of the object.
% (kinematically feasible? successful?)
% the grasp success (object is between gripper)
% and the pose of the object relative to the gripper. 
Furthermore, we log the pose difference after grasping and lifting the object (chosen grasp pose relative to the gripper TCP) and placement (current object pose relative to defined placement pose). We consider an object as successfully lifted / grasped or placed precisely enough if the distance between the pose representatives based on~\cite{Bregier_PoseDistance} is less than 0.1 times the diameter of the smallest bounding sphere of the object. This is analogous to the metric used for object pose estimation in computer vision proposed by Br\'{e}gier et al.~\cite{Bregier_PoseDistance,Bregier_SymmetryAwareEvaluation} and allows to properly consider all possible kinds of object symmetry.
Furthermore, we log for each grasp pose whether an entanglement with other objects in the scene occurred.

% We consider an object as placed correctly if the distance between the pose representatives of the define disposal (endpose) pose and the actual pose is less than 0.1 times the diameter of the object, which is the same metric used for (evaluating the) object pose estimation.

% Loop over all objects
% Loop over all grasp poses
% Log:
% - ...
% - unrest metric
% - entanglements

%%% Symmetries...
Since the grasp poses are defined relative to the object coordinate system, pose ambiguities due to object symmetries result in convergence issues during neural notwork training. To avoid this, we introduce a unique object pose definition.
% We therefore unify the object pose (make object pose unique) in the simulation.
For discrete symmetries, we ensure to pick the pose where the $z$-component of a non-symmetry axis ($x$- or $y$-axis) is maximal (assuming the axis of symmetry is the $z$-axis). For continuous symmetries, we maximize the $z$-component of a non-symmetry axis by rotating around the axis of symmetry.

% also has to be done in inference mode for pose estimates... (in general)

%%%%%%%%%%%%%%%%%%%%%%%%%%%%%%%%%%%%%%%%%%%%%%%%%%%%%%%%%%%%%%%%%%%%%%%%%%%%%%%%
%%%%%%%%%%%%%%%%%%%%%%%%%%%%%%%%%%%%%%%%%%%%%%%%%%%%%%%%%%%%%%%%%%%%%%%%%%%%%%%%
%%%%%%%%%%%%%%%%%%%%%%%%%%%%%%%%%%%%%%%%%%%%%%%%%%%%%%%%%%%%%%%%%%%%%%%%%%%%%%%%
%\subsubsection{Defining the Graspability}
%\subsection{Defining the Graspability}
\subsection{Graspability Metrics}
\begin{figure*}[thpb]
\centering
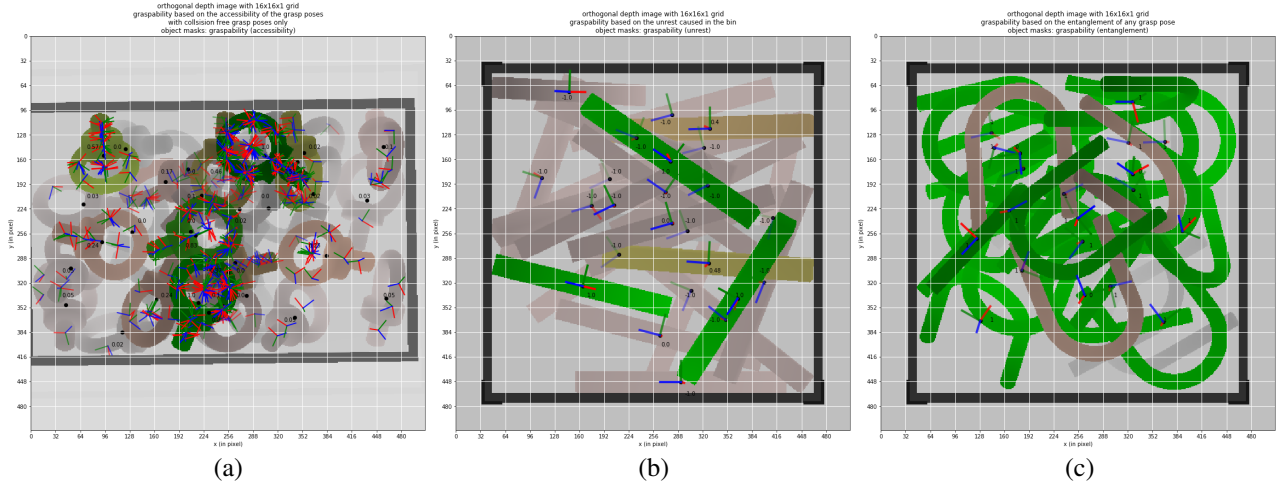
\caption{Exemplary ground truth samples: (a) Graspability $g_\mathrm{a} \in [0,1]$ of the IPARingScrew based on the accessibility of the grasp poses (each collision-free grasp pose is indicated by a small coordinate system). (b) Graspability $g_\mathrm{u} \in [0,1]$ of the IPABar based on the unrest caused in the bin
%/scene
during grasping. (c) Graspability $g_\mathrm{e} \in \{0,1\}$ of the IPAUBolt based on entanglements with other objects after grasping. Objects shown in green have a high rating for tangibility. The more difficult an object is to grasp, the more the colour changes from green to yellow to red with increasing transparency.
%(in the bin/scene)
}
\label{fig:Graspability}
\end{figure*}

Using the results from all executed grasps, we define instance based metrics to assess
% evaluate
the graspability of each object in the scene.
%%%%%%%%%%%%%%%%%
% accessibility %
%%%%%%%%%%%%%%%%%
The graspability of an object based on the accessibility of the grasp poses $g_\mathrm{a} \in [0,1]$ is defined as the ratio between the number of collision-free grasps and
%$\frac{1}{4}$ one quarter of
the number of total grasps $J$.
%\begin{equation}
%g_\mathrm{a}=\sum_J 1/4 n_gp
%\end{equation}
Fig.~\ref{fig:Graspability}~(a) shows the ground truth  labels $g_\mathrm{a}$ for the IPARingScrew. Some fully visible objects (which are easy to localize) cannot be grasped because other objects are in the way, demonstrating that visibility and graspability of objects are not fully correlated.

%%%%%%%%%%
% unrest %
%%%%%%%%%%
For a removal that is gentle on the components
%(für eine bauteilschonende Entnahme)
and to avoid movements of the grasped object relative to the gripper due to overlapping objects, entanglements, or jams preventing a precise object placement, we log the movement in terms of position $\mat{x} \in \NewR^3$ of all other objects in the scene before grasping ($t_0$) and after lifting ($t_1$). This information is used
%(sum of the Euclidean distances)
to define
% a metric ($g_\mathrm{u}$)
the graspability of object $k$ based on the unrest caused in the bin
%/scene
%by/
during grasping ("mikado metric")
\begin{equation}
g_{\mathrm{u},k}=1-\min \Bigg(\sum_{n=1,n\neq k}^{N}\lVert\mat{x}_{n,t_0}-\mat{x}_{n,t_1}\lVert,1\Bigg)
\end{equation}
with $\lVert.\rVert$ being the $L^2$ norm and $N$ being the number of objects in the scene without the picked object $k$~\cite{MA_Moosmann}.
Fig.~\ref{fig:Graspability}~(b) shows exemplary ground truth $g_\mathrm{u} \in [0,1]$ labels for the IPABar object. It can be seen that the objects at the top of the bin have a high graspability value regarding the unrest.

%%%%%%%%%%%%%%%%
% entanglement %
%%%%%%%%%%%%%%%%
The graspability of object $k$ based on the entanglement with other objects
%(of any grasp pose)
is
\begin{equation}
g_{\mathrm{e},k}=\begin{cases}
    0,~\text{if an entanglement occurred for any grasp pose}\\
    1,~\text{otherwise}.
\end{cases}
\end{equation}
% $g_\mathrm{e} \in {0,1}$
Fig.~\ref{fig:Graspability}~(c) shows the ground truth $g_\mathrm{e}$ labels for the IPAUBolt. A goal for the robot is to avoid picking objects which can potentially entangle.

%%%%%%%%%%%%%%%%%%%%%%%%%%%%%%%%%%%%%%%%%%%%%%%%%%%%%%%%%%%%%%%%%%%%%%%%%%%%%%%%
%%%%%%%%%%%%%%%%%%%%%%%%%%%%%%%%%%%%%%%%%%%%%%%%%%%%%%%%%%%%%%%%%%%%%%%%%%%%%%%%
%%%%%%%%%%%%%%%%%%%%%%%%%%%%%%%%%%%%%%%%%%%%%%%%%%%%%%%%%%%%%%%%%%%%%%%%%%%%%%%%
\subsection{Parameterization of the Output}
Similar to~\cite{OP-Net}, we introduce a spatial discretization of the 3D scene into $S \times S$ volume elements (see white grid in Fig.~\ref{fig:Graspability}) and solve a regression problem locally, i.e., individually for each volume element. Each volume element comprises an $(11+J)$-dimensional vector containing the probability $p$, visibility $v$, positions $x$, $y$, $z$, Euler angles $\varphi_1$, $\varphi_2$, $\varphi_3$, graspabilities $g_\mathrm{a}$, $g_\mathrm{u}$, $g_\mathrm{e}$ of the object, and a $J$-dimensional vector with a success label $s \in \{0,1\}$ for each grasp pose $G$ for the considered task (grasping, precise placement).
% relative to the camera coordinate system
% grasp pose success
For the ground truth generation, the objects are assigned to the volume element
%$i$
which contains the origin of the object coordinate system.
%, i.e. $p_i=1$.
In case multiple objects fall into the same volume element, we assign the object with the highest visibility $v$ as ground truth.
%If multiple origins fall into the same volume element, the object with higher visibility $v$ is assigned as ground truth.
All volume elements not containing an object are filled with a zero vector. The output of the network is a $S \times S \times (11+J)$ tensor as depicted in Fig.~\ref{fig:Overview} (d).

%%%%%%%%%%%%%%%%%%%%%%%%%%%%%%%%%%%%%%%%%%%%%%%%%%%%%%%%%%%%%%%%%%%%%%%%%%%%%%%%
%%%%%%%%%%%%%%%%%%%%%%%%%%%%%%%%%%%%%%%%%%%%%%%%%%%%%%%%%%%%%%%%%%%%%%%%%%%%%%%%
%%%%%%%%%%%%%%%%%%%%%%%%%%%%%%%%%%%%%%%%%%%%%%%%%%%%%%%%%%%%%%%%%%%%%%%%%%%%%%%%
\subsection{Loss Function}

To train the network, the multi-task
% (multi-part)
loss function
%\begin{equation}
%\mathcal{L} = \sum_{i=1}^{S^2} \bigg( \lambda_1(p_i - \hat{p}_i)^2+\Big[ \lambda_2(v_i - \hat{v}_i)^2 + %\lambda_3\big( \mathcal{L}_\mathrm{pos} + \lambda_4 \mathcal{L}_\mathrm{ori} \big) \Big] p_i \bigg)
%\end{equation}
\begin{equation}
\mathcal{L} = \sum_{i=1}^{S^2} \bigg( \lambda_1\mathcal{L}_\mathrm{p}+\Big[ \lambda_2\mathcal{L}_\mathrm{v} + \lambda_3\mathcal{L}_\mathrm{pose}+\lambda_5\mathcal{L}_\mathrm{g} + \lambda_6\mathcal{L}_\mathrm{gp_s} \Big] p_i \bigg)
\end{equation}
is optimized. The $\lambda$-factors
% $\lambda_1$, $\lambda_2$, $\lambda_3$, $\lambda_4$, $\lambda_5$, and $\lambda_6$
are manually tuned weights for the different loss terms. While $\lambda_1=0.1$, $\lambda_2=0.1$, $\lambda_4=1$, $\lambda_5=1$, and $\lambda_6=1/J$ are constant, $\lambda_3=(g_\mathrm{a}+g_\mathrm{u}+g_\mathrm{e})^3$ is a function of the ground truth graspabilities $g_\mathrm{a}, g_\mathrm{u}, g_\mathrm{e}$ to make the network focus on the relevant objects for grasping.

For the loss of the pose
\begin{equation}
\mathcal{L}_\mathrm{pose}=\mathcal{L}_\mathrm{pos} + \lambda_4 \mathcal{L}_\mathrm{ori}
\end{equation}
we use
\begin{equation}
\mathcal{L}_\mathrm{pos} = \lVert\mat{x}-\hat{\mat{x}} \rVert^2
\end{equation}
with $\mat{x}=[x, y, z]^\top$ and
\begin{equation}
\mathcal{L}_\mathrm{ori} = \lVert\bm{\varphi}-\hat{\bm{\varphi}}\rVert^2
\end{equation}
with $\bm{\varphi}=[\varphi_1, \varphi_2, \varphi_3]^\top$
%and $\varphi_j \in [0,2\pi/k_j)$ for $j=1,2,3$
and $\varphi_1,\varphi_2 \in [0,2\pi)$ and $\varphi_3 \in [0,2\pi/k)$,
where $k \in \mathbb{N}$ represents the order of the cyclic symmetry.
% (and ground truth angles $\varphi_i \in [0,2\pi)$)

To stabilize the training, the position $\mat{x} \in \NewR^3$ of the object is estimated relative to the volume element, i.e., $x, y, z \in (0,1)$, while $z$ is the position between the near and far clipping plane of the 3D sensor, and the angles are bounded, i.e.,
$\varphi_1,\varphi_2,\varphi_3$
%$\varphi_j \in [0,2\pi/k_j)$
are mapped to $[0,1)$. For objects with a revolution symmetry, we omit the respective output feature-map.

We use the binary cross-entropy loss for the probability channel, the visibility channel, the three graspability channels, and the grasp pose result channels to compute $\mathcal{L}_\mathrm{p}$, $\mathcal{L}_\mathrm{v}$, $\mathcal{L}_\mathrm{g}$, and $\mathcal{L}_\mathrm{gp_s}$, respectively, while only backpropagating the loss for the elements that contain ground truth
%by multiplying with $p_i \in \{0,1\}$.
by multiplying each channel element-wise with the ground truth probability channel.
%With
%($p_i \in \{0,1\}$ and $\hat{p}_i \in [0,1]$ where a hat indicates estimates of the network).

%L g_a
%L g_u
%L g_e

%\begin{equation}
%\mathcal{L}_{gp_s}=\sum_{j=1}^{J} \lVert r_j - \hat{r}_j \lVert^2
%\end{equation}

% squared L2-norm / BCE for visibility, graspability, ...

%%%%%%%%%%%%%%%%%%%%%%%%%%%%%%%%%%%%%%%%%%%%%%%%%%%%%%%%%%%%%%%%%%%%%%%%%%%%%%%%
%%%%%%%%%%%%%%%%%%%%%%%%%%%%%%%%%%%%%%%%%%%%%%%%%%%%%%%%%%%%%%%%%%%%%%%%%%%%%%%%
%%%%%%%%%%%%%%%%%%%%%%%%%%%%%%%%%%%%%%%%%%%%%%%%%%%%%%%%%%%%%%%%%%%%%%%%%%%%%%%%
\subsection{Network Architecture}
The input of our model is a single normalized depth image
%in perspective projection.
which is
%The data is
processed with a fully convolutional architecture and mapped to a 3D output tensor as shown in Fig.~\ref{fig:Overview}~(d). In the experiments, we use an input resolution of $128 \times 128$ pixel,
a DenseNet-BC~\cite{DenseNet} with 40 layers and a growth rate of 50, which represents the number of feature-maps being added per layer, and $S=16$. We choose a DenseNet-BC because it promotes gradient propagation by introducing direct connections between any two layers with the same feature-map size and has a high parameter efficiency.
% the same manually optimized weighting factors $\lambda_1=1/S$, $\lambda_2=5$, and $\lambda_3=2v^3$.
The
%fully convolutional
network architecture consists of four dense blocks and downsampling is performed three times via $2 \times 2$ average pooling to reduce the size of the feature-maps from $128 \times 128$ to $16 \times 16$ and preserve the spatial information.
ReLU activation functions are employed in the dense blocks and sigmoid functions for
%(all channels of)
the 3D output tensor.
With this architecture, forward passes
%through the \mbox{PQ-Net}
are performed with a frame rate of 92~fps on a Nvidia Tesla V100.
%\mbox{PQ-Net} has a
%This results in a
%%% average...
%frame rate of 92~fps
% 11 ms per depth image
%on a Nvidia Tesla V100 for the feed-forward.

%%%%%%%%%%%%%%%%%%%%%%%%%%%%%%%%%%%%%%%%%%%%%%%%%%%%%%%%%%%%%%%%%%%%%%%%%%%%%%%%
%%%%%%%%%%%%%%%%%%%%%%%%%%%%%%%%%%%%%%%%%%%%%%%%%%%%%%%%%%%%%%%%%%%%%%%%%%%%%%%%
%%%%%%%%%%%%%%%%%%%%%%%%%%%%%%%%%%%%%%%%%%%%%%%%%%%%%%%%%%%%%%%%%%%%%%%%%%%%%%%%
\subsection{Training}
\label{sec:training}
During training, the error of the entire probability channel and the error of the remaining elements from the 3D output tensor that contain ground truth are backpropagated.
% "Trick" (annotating all scenes is not needed)
Since annotating the grasps is time-consuming, we do not annotate the whole training dataset
%(Instead) During training, we
and only backpropagate
%the grasping error
$\mathcal{L}_\mathrm{g}$ and $\mathcal{L}_\mathrm{gp_s}$ for the samples, where ground truth annotations are available.
In our experiments, we only annotate 100 out of 750 cycles from the Fraunhofer IPA~\cite{FraunhoferIPABinPickingDataset} and our newly provided dataset, resulting in 100 uniformly distributed samples over different fill levels of the bin.
% annotation of the scenes with success labels for each grasp pose is expensive (time)
% --> we only require 100 annotated cycles

%\subsection{Data Augmentation}
%rotate images
%mirror images
We augment the training data by rotating around the $z$-axis of the camera coordinate system and mirroring the images if the object is symmetric with respect to a plane while adjusting the ground truth pose annotations accordingly. To not lose the information of the robot placement relative to the bin,
%/scene)
we only backpropagate $\mathcal{L}_\mathrm{gp_s}$ for the non-augmented samples.

%\subsection{Sim-to-Real Transfer}
For a robust Sim-to-Real Transfer, we use domain randomization~\cite{Domain_Randomization}. To allow the model generalizing on real-world data, we apply different augmentations with varying intensity to the rendered training images, e.g., adding noise, blurring, elastic transformations, dropout, etc. This allows \mbox{PQ-Net} generalizing to different 3D sensor technologies.
%(as demonstrated in~\cite{OP-Net}).

% Optimization
We use the Adam optimizer with an initial learning rate of 0.01, monitor the validation loss, reduce the learning rate by a factor of 10 if the loss did not improve for three epochs, and train the network for about 50 epochs on the synthetic data.
% generated by the physics simulation.

%%%%%%%%%%%%%%%%%%%%%%%%%%%%%%%%%%%%%%%%%%%%%%%%%%%%%%%%%%%%%%%%%%%%%%%%%%%%%%%%
%%%%%%%%%%%%%%%%%%%%%%%%%%%%%%%%%%%%%%%%%%%%%%%%%%%%%%%%%%%%%%%%%%%%%%%%%%%%%%%%
%%%%%%%%%%%%%%%%%%%%%%%%%%%%%%%%%%%%%%%%%%%%%%%%%%%%%%%%%%%%%%%%%%%%%%%%%%%%%%%%
%\subsection{Grasping Policy}
\subsection{Policy}
% learning robust grasping policies
Based on a single
% input
depth image $I$ with global scene information, the neural network $f$ with weights $\theta$ outputs a 3D tensor $\hat{T}$. Our policy $\pi$ uses the network output to select
% the most robust /
the highest quality grasp weighted with $\hat{p}$, $\hat{v}$, $\hat{g_\mathrm{a}}$, $\hat{g_\mathrm{u}}$, and $\hat{g_\mathrm{e}}$ from all $S^2$ volume elements for execution
\begin{equation}
\pi(f_\theta(I)) = \argmax_{i,j}(\hat{s}_{j,i} \cdot \hat{p}_i \cdot \hat{v}_i \cdot \hat{g_\mathrm{a}}_i \cdot \hat{g_\mathrm{u}}_i \cdot \hat{g_\mathrm{e}}_i)
\end{equation}
with $i=1,...,S^2$, $j=1,...,J$ with $J$ being the number of predefined grasp poses, and $\hat{s}_{j,i}$ being the success estimate of grasp pose $G_j$ at volume element $i$
%and $\hat{\mat{s}}$ comprising a success estimate
%the estimated result
%for each grasp pose
for the considered task (grasping, precise placement).

% "executing the highest quality grasp" % (paper Dex-Net)
% by selection the argmax for the grasp pose
% argmax of the network output
% Grasp selection: argmax

% (ToDo: How to write this formally correct?)

%%%%%%%%%%%%%%%%%%%%%%%%%%%%%%%%%%%%%%%%%%%%%%%%%%%%%%%%%%%%%%%%%%%%%%%%%%%%%%%%
%%%%%%%%%%%%%%%%%%%%%%%%%%%%%%%%%%%%%%%%%%%%%%%%%%%%%%%%%%%%%%%%%%%%%%%%%%%%%%%%
%%%%%%%%%%%%%%%%%%%%%%%%%%%%%%%%%%%%%%%%%%%%%%%%%%%%%%%%%%%%%%%%%%%%%%%%%%%%%%%%
%\section{Experiments}
\section{Experimental Evaluation}
\label{sec:evaluation}

In this paper, we focus on parallel jaw grippers
% ((only))
using a RG2 gripper~\cite{RG2_gripper}. Given a proper physics simulation, our approach can easily be transferred to other gripper types.

\subsection{Sim-to-Real Transfer}

% collision-free reachability
% threshold vis =0.5?
% threshold success=0.5?
\begin{table*}[h]$ $
\begin{tiny}
\caption{Prediction of collision-free reachability of grasp poses of our approach on real-world / noisy data for different objects with different kinds of object symmetry form the Sil\'{e}ane~\cite{Bregier_SymmetryAwareEvaluation} and Fraunhofer IPA~\cite{FraunhoferIPABinPickingDataset} datasets.}
\label{table:collision_free_reachability}
\begin{center}
\begin{tabular}{|l||c|c|c|c|c|c|c|}
\hline
object & SileaneBunny & SileaneCandlestick & SileanePepper & SileaneGear & SileaneTLess20 & IPARingScrew & IPAGearShaft \\
\hline
object symmetry based on~\cite{Bregier_PoseDistance,Bregier_SymmetryAwareEvaluation} & no proper & revolution & revolution & revolution & cyclic & cyclic & revolution \\
 & symmetry & & & & (order 2) & (order 2) & \\
\hline
% Grasping:
\hline
PQ-Net success rate of policy & 0.98 & 0.97 & 0.98 & 0.99 & 0.98 & 0.98 & 0.99 \\
\hline
PQ-Net precision (all grasp poses) & 0.57 & 0.70 & 0.71 & 0.73 & 0.77 & 0.75 & 0.83 \\
\hline
PQ-Net recall (all grasp poses) & 0.52 & 0.67 & 0.66 & 0.64 & 0.65 & 0.45 & 0.65 \\
\hline
% OPE:
\hline
PQ-Net success rate OPE for chosen object & 0.89 & 0.92 & 0.95 & 0.98 & 0.98 & 0.98 & 0.99 \\
\hline
PQ-Net success rate OPE for chosen object with ICP & 0.91 & 0.93 & 0.95 & 0.99 & 0.99 & 0.99 & 0.99 \\
\hline
PQ-Net AP (OPE) whole scene & 0.86 & 0.88 & 0.92 & 0.74 & 0.82 & 0.86 & 0.98 \\
\hline
OP-Net~\cite{OP-Net} AP (OPE) whole scene & 0.92 & 0.95 & 0.98 & 0.82 & 0.85 & 0.88 & 0.99 \\
\hline
\end{tabular}
\end{center}
\end{tiny}
\end{table*}
% (threshold=0.5)
% ADD / ADI

% ((For showing the feasibility of our approach))
% successful transfer of the policy to real-world data
For demonstrating a robust transfer of
% our policy / appraoch /
\mbox{PQ-Net} to the real world, we extend the Sil\'{e}ane~\cite{Bregier_SymmetryAwareEvaluation} and Fraunhofer IPA~\cite{FraunhoferIPABinPickingDataset} datasets with annotations for the collision-free reachability of approximately 500 densely sampled grasp poses (examples see Fig.~\ref{fig:GraspPoses}). In Table~\ref{table:collision_free_reachability} we report the success rate of our grasping policy and the precision and recall over all
%(possible) do not use
grasp poses in the scene.
% (threshold visibility=0.5?!)
%
% (Conclusions drawn? Works well.)
% indicating very high success rates even on...
%
Applying randomizations on the synthetic images during training (see Section~\ref{sec:training}) allows \mbox{PQ-Net} providing robust pose estimates and very high success rates of the policy on real-world data recorded with different 3D sensors.
% different real-world sensor data.

Table~\ref{table:collision_free_reachability} gives the average precision (AP) results for the object pose estimation based on the metric provided by Br\'{e}gier et al.~\cite{Bregier_PoseDistance, Bregier_SymmetryAwareEvaluation} from OP-Net~\cite{OP-Net} with exactly the same depth image resolution, network architecture, and output discretization as used for \mbox{PQ-Net}. Even with a success estimate for approximately 500 grasp poses together with the graspability $g_\mathrm{a}$, \mbox{PQ-Net} only loses very few points in terms of AP.
% For comparable AP results with OP-Net, the network capacity has to be increased.
%
In addition to the datasets, videos of real-world experiments are available at %\url{https://owncloud.fraunhofer.de/index.php/s/975EH0WxWYG0noY}.
\url{http://www.bin-picking.ai/en/dataset.html}.

%Sim-to-Real Transfer: Annotate the noisy data from the Sil\'{e}ane dataset and real-world data from the "Fraunhofer IPA Bin-Picking dataset" via simulation and report result on that (precision, recall, ...)
% annotation of the collision-free reachability is very fast

% Using the estimation for the collision-free grasps together with the estimation of the graspability based on the accessibility --> can be used as a massive speedup for the heuristic search method proposed in [FPS]

% We add additional objects since IPARingScrew and IPAGearShaft might start moving after setting to dynamic --> no nice annotations/labels...

% Evaluate the grasping policy

%Success/result levels:
%- No collision at grasp pose
%- Kinematically feasible (collisison-free reachability)
%- Success grasping (no entanglement; successful grasps with entanglements are excluded)
%- Placeable (kinematically)
%- Successful placement (based on the pose distance based on metric by Br\'{e}gier et al.); Definied/precise placement;

%%%%%%%%%%%%%%%%%%
%%% Comparison %%% Comparison of PQ-Net with GG-CNN, bp3, ... in simulation
%%%%%%%%%%%%%%%%%%
%\subsection{Grasping and Precise Placement}
%\subsection{Comparison/Benchmarking in Simulation}
\subsection{Benchmarking in Simulation}

% Structure:
% 1. Comparison
% 2. Challenges
% - IPABar
% - IPAUBolt
% 3. GG-CNN (learning-based approach)
% 4. Analytical approach
% 5. PQ-Net (ours)

For evaluation, we compare the performance of three approaches
%GG-CNN~\cite{GG-CNN, GG-CNN_2} and \mbox{PQ-Net}
%of two/three different approaches
in simulation on two very challenging objects. Each approach gets to observe the same 250 scenes for both the IPABar (see Fig.~\ref{fig:Motivation} left) and IPAUBolt (see Fig.~\ref{fig:Motivation} right), respectively, and executes one grasp per scene to ensure a comparison under the same conditions (the approaches face the exactly same scenarios). Table~\ref{table:grasping_and_precise_placement} reports the success rates for each method.

% Clarify: Why are IPABar and IPAUBolt objects challenging?
%%%%%%%%%%%%%%
%%% IPABar %%%
%%%%%%%%%%%%%%
A robust picking of the IPABar objects from a cluttered bin is challenging because they require a defined picking order. If an occluded object is chosen for grasping, the grasp trial might fail completely or the object might move relative to the gripper which hinders a precise placement of the object.
Therefore, the right object and grasp pose has to be chosen from the highly cluttered scene. This is especially important for friction grasps (force closure) because for a form closure it is unlikely that object moves relative to the gripper.

% force closure (less stable than form closure)
% form closure (easier) --> it is less likely that object moves relative to the gripper having a good form closure

% IPABar cannot be place precisely if the wrong object and/or wrong grasp pose is chosen in the highly cluttered scene...

%%%%%%%%%%%%%%%%
%%% IPAUBolt %%%
%%%%%%%%%%%%%%%%
The IPAUBolt is challenging because it can potentially entangle with other objects in the bin.
In case a wrong object and grasp pose for picking is chosen,
%Choosing a wrong object and grasp pose for picking, 
the robot might lift multiple objects resulting in failed grasp because no collision-free placement of the object is possible.

%%%%%%%%%%%%%%
%%% GG-CNN %%%
%%%%%%%%%%%%%%
GG-CNN~\cite{GG-CNN, GG-CNN_2}
%(learning-based approach)
and other model-free approaches~\cite{Dex-Net_1.0,Dex-Net_2.0,Dex-Net_3.0,JosephRedmon.2015,IanLenz.2015,Google_2016_Learning_Hand-Eye_Coordination,Google_2018_Learning_Hand-Eye_Coordination,Morrison_MultiViewPicking,QT_Opt} focus on generalization performance, use top-down grasps for pick-and-drop of the objects, and cannot solve precise \mbox{pick-and-place} tasks.
% (No precise placement possible).
% with GG-CNN no precise placement possible
To compare our approach with these methods, we use the logged
%grasp pose success labels for
grasping success labels for training.
% we define result labels for each grasp pose as successful if an object is between the gripper fingers after moving up / lifting.
Table~\ref{table:grasping_and_precise_placement} reports the success rates for each approach.
% Similar to other model-free approaches [XX,XX,XX,...] GG-CNN performs top-down grasps.
Our approach outperforms GG-CNN
%~\cite{GG-CNN, GG-CNN_2}
because of operating in 6D and being specifically configured to the object.
% (while GG-CNN focuses on generalization performance)
%/ generalization to novel objects).
Furthermore, GG-CNN collides with the bin when attempting to grasp object close to the border due to the limited flexibility in the gripper orientation.

% physical grasping
%\begin{table}[h]
\begin{table*}[hbtp]$ $
\caption{Comparison of the performance of \mbox{PQ-Net} with other approaches and performance evaluation.
%Best results in bold.
}
\label{table:grasping_and_precise_placement}
\begin{center}
\begin{tabular}{|l||c|c|}
\hline
object & IPABar & IPAUBolt \\
\hline
object symmetry based on~\cite{Bregier_PoseDistance,Bregier_SymmetryAwareEvaluation} & finite non trivial & cyclic (order 2) \\
\hline
\hline
GG-CNN~\cite{GG-CNN, GG-CNN_2} success rate for grasping & 0.78 & 0.67 \\
\hline
GG-CNN~\cite{GG-CNN, GG-CNN_2} success rate for grasping without bin & 0.81 & 0.72 \\
%\hline
%success rate (grasping) analytical approach~\cite{FPS_GPC_1} & 0.84 & 0.83 \\
\hline
PQ-Net (ours) success rate for grasping & 0.99 & 0.87 \\
\hline
PQ-Net (ours) precision (all grasp poses) & 0.63 & 0.59 \\
\hline
PQ-Net (ours) recall (all grasp poses) & 0.65 & 0.57 \\
\hline
% for grasping only OPE is not important
\hline
analytical approach~\cite{FPS_GPC_1} success rate for precise placement & 0.85 & 0.80 \\
\hline
PQ-Net (ours) success rate for precise placement & 0.89 & 0.81 \\
\hline
%\hline
PQ-Net (ours) success rate OPE & 0.96 & 0.85 \\
%\hline
%PQ-Net (ours) success rate OPE with ICP & 0.96 & 0.85 \\
%\hline
%Ours AP (OPE) whole scene & 0.XX & 0.XX \\
\hline
PQ-Net (ours) precision (all grasp poses) & 0.57 & 0.66 \\
\hline
PQ-Net (ours) recall (all grasp poses) & 0.52 & 0.60 \\
\hline
\end{tabular}
\end{center}
\end{table*}

%%% (ADD/ADI)

%%%%%%%%%%%%%%%%%%%%%%%%%%%%%%%%%%%%%%%%%%%%%%%%%%
%%% classical/analytical approaches (e.g. bp3) %%%
%%%%%%%%%%%%%%%%%%%%%%%%%%%%%%%%%%%%%%%%%%%%%%%%%%

% bp3: “GPC versucht möglichst schnell eine greifbare Lösung zu finden”
% GPC: prüfen, welches Bauteil kollisionsfrei gegriffen werden kann; Ziel: kollisionsfrei Bahn finden

%Fails bp3 (analytical approach) (reasons for outperforming bp3):
%- Verhakungen (Entanglements)
%- Verklemmungen (jams)
%- ...

%The analytical approach tries to find a collision-free robot robot configuration and path (e.g. as fast as possible using heuristic search) and does not give an estimate on the actual physical outcome of the grasp. 
The analytical approach considers the collision-free reachability and kinematical feasibility of the grasp pose and the path only and does not give an estimate on the actual physical outcome of the grasp, e.g., whether the object might move relative to the gripper in the given scenario,
%movement of the object relative to the gripper (because only checks collision-free reachability)
%Furthermore, it do not consider 
jamming,
%of the object
or entanglements with other objects in the bin.

%%%%%%%%%%%%%%
%%% PQ-Net %%%
%%%%%%%%%%%%%%
With our simulation-driven and learning-based approach, we let our system autonomously learn how to localize and grasp the objects
%in the bin/scene
and transfer the automatically gained experience from the simulation to the real world without any time-consuming object-specific manual tuning.

%%%%%%%%%%%%%%%%%%
%%% Add on ... %%%
%%%%%%%%%%%%%%%%%%
\iffalse
Since the success labels from the physics simulation are annotated with respect to the ground truth pose, the accuracy of the pose estimation is very important.
%because we use the annotations from the physics simulation with respect to the ground truth pose, we also give the success rate of the OPE of our approach in table XX.
% --> pose estimates are already very good
% --> run ICP on the object that will be grasped next (chosen object)
Tables~\ref{table:collision_free_reachability} and \ref{table:grasping_and_precise_placement} show the success rates for the OPE based on the metric by Br\'{e}gier et al.~\cite{Bregier_PoseDistance,Bregier_SymmetryAwareEvaluation} for the chosen object
%((AP values))
%after applying ICP refinement.
with and without ICP refinement. Due to the high success rates, the robot can reliably grasp the object for precise placement.
\fi

%%%%%%%%%%%%%%%%%%%%%%%%%%%%%%%%%%%%%%%%%%%%%%%%%%%%%%%%%%%%%%%%%%%%%%%%%%%%%%%%
%%%%%%%%%%%%%%%%%%%%%%%%%%%%%%%%%%%%%%%%%%%%%%%%%%%%%%%%%%%%%%%%%%%%%%%%%%%%%%%%
%%%%%%%%%%%%%%%%%%%%%%%%%%%%%%%%%%%%%%%%%%%%%%%%%%%%%%%%%%%%%%%%%%%%%%%%%%%%%%%%
\section{Discussion}
\label{sec:discussion}
In the following, we summarize strengths and discuss limitations of our approach.

%%%%%%%%%%%%%%%%%%%%%%%%%%%%%%%%%%%%%%%%%%%%%%%%%%%%%%%%%%%%%%%%%%%%%%%%%%%%%%%%
%%%%%%%%%%%%%%%%%%%%%%%%%%%%%%%%%%%%%%%%%%%%%%%%%%%%%%%%%%%%%%%%%%%%%%%%%%%%%%%%
%%%%%%%%%%%%%%%%%%%%%%%%%%%%%%%%%%%%%%%%%%%%%%%%%%%%%%%%%%%%%%%%%%%%%%%%%%%%%%%%
\subsection{Strengths}

% - Global context / global reasoning --> near optimal decisions; global reasoning --> better results (than GG-CNN, analytical approach?, ...)

\mbox{PQ-Net} gets to observe the whole depth image and selects highly robust grasps on a global level because of not looking at local patches of the image only.
Our approach can operate in a closed-loop fashion with 92~fps for the forward pass (for OPE and grasp planning) and is therefore suitable for grasping in non-static environments.
% can be used for closed-loop grasping
Our graspability metrics allow a gentle removal of the components and avoiding to grasp entangling objects.
% - Unrest metric --> Materialschonende Entnaheme möglich
Furthermore, our approach can be extended to prioritizing grasp poses which allow a precise placement without re-grasping (e.g., important for objects without symmetries) which allows to reduce cycle times.
% prioritize objects which can be picked without the need for re-grasping for a precise placement
\mbox{PQ-Net} provides robust estimates on real-world data independent of the actual 3D technology being used
% (domain randomization)
and does not require any human labeled data or grasping trials on the real-world
%physical
system, facilitating scalability. Furthermore, it
% operate on real data without any fine-tuning (Zero-Shot Transfer); independent of the actual 3D technology being used (domain randomization)
%- contrary to approaches trained in the real world [Google, or using human labeled datasets [GG-CNN],
% --> limited in the scalability to different hardware setups...
%autonomously/
automatically configures
%(for a given hardware setup)
for new object geometries using simulation and machine learning for precise
%object placement
\mbox{pick-and-place} tasks in highly cluttered scenes by providing an object model only.
% also assign:  Lenz et al. (Two-Stage System), Redmon et al., Supersizing self-supervision(?)
% S. Kumra and C. Kanan. Robotic Grasp Detection using Deep Convolutional Neural Networks. (IROS 2017) (from paper GG-CNN)
% approaches requiring human label or grasp trials on the physical system are ...
Our approach properly considers all possible kinds of object symmetries during data generation in the physics simulation and in the loss function for the regression of the angles.
Furthermore, our provided simulated scenes can be used to benchmark further approaches.

% One will notice that the hardware setup is not optimal... (when for some objects no grasp pose works without collisions)

% outlook / future work
% Advantage our approach: Grasp selection based on fastest/easiest motion possible

%%%%%%%%%%%%%%%%%%%%%%%%%%%%%%%%%%%%%%%%%%%%%%%%%%%%%%%%%%%%%%%%%%%%%%%%%%%%%%%%
%%%%%%%%%%%%%%%%%%%%%%%%%%%%%%%%%%%%%%%%%%%%%%%%%%%%%%%%%%%%%%%%%%%%%%%%%%%%%%%%
%%%%%%%%%%%%%%%%%%%%%%%%%%%%%%%%%%%%%%%%%%%%%%%%%%%%%%%%%%%%%%%%%%%%%%%%%%%%%%%%
\subsection{Limitations} % Weaknesses
\mbox{PQ-Net} is a model-based approach and, therefore, does not generalize to unseen objects. Instead, it configures for novel objects without any human intervention.
%/involvement.
%(but automatic configuration without manual tuning)
The approach requires a large dataset to train on, where the process of
%- Long time for
data generation (grasp execution) can take long, especially, when increasing the number of grasp poses. Still the process for data generation can run much faster than real time and can easily be parallelized across multiple machines.
Furthermore, our method cannot do pre-grasp manipulations on the objects to change their poses
% of the object to be able to 
in order to more robustly grasp them.
While \mbox{PQ-Net} can avoid entangled object situations,
% (similar to [IROS2019, MAM])
we do not propose a solution to unhook very complex object geometries, for which no general solution has been proposed so far~\cite{Potentially_Tangled_Objects,paper_MAM}.
% - avoid entanglements, jams (Verklemmungen), ... (but do not provide a solution to solve these)

% Further:
% - balancing many loss terms...
% - (No experiments: completely empty the bin)
% - Requires a dense discretization of the grasp poses to be able to fully empty the bin; could give a success prediction for continuous areas and use a different approach to find a grasp pose in that area quickly

%%%%%%%%%%%%%%%%%%%%%%%%%%%%%%%%%%%%%%%%%%%%%%%%%%%%%%%%%%%%%%%%%%%%%%%%%%%%%%%%
%%%%%%%%%%%%%%%%%%%%%%%%%%%%%%%%%%%%%%%%%%%%%%%%%%%%%%%%%%%%%%%%%%%%%%%%%%%%%%%%
%%%%%%%%%%%%%%%%%%%%%%%%%%%%%%%%%%%%%%%%%%%%%%%%%%%%%%%%%%%%%%%%%%%%%%%%%%%%%%%%
\section{Conclusion and Future Work}

In this paper, we proposed a novel learning-based approach
%for object pose estimation and grasp planning
for grasping in highly cluttered scenes and precise object placement based on depth images.
% / precise \mbox{pick-and-place} tasks.
Our approach outputs the 6D object poses together with a graspability and quality estimate for each automatically generated grasp pose for multiple objects simultaneously in a single forward pass in a joint framework (running at 92~fps).
%
%Based on a dense discretization of possible grasp poses, we provide a learning-based approach to select a highly robust grasp in highly cluttered scenes with heavy occlusions.
% Based on a dense discretization of possible grasp poses, ...
All densely discretized and automatically generated grasp poses are executed in a physics simulation and the gained experience is transferred from simulation
%(physically simulated grasps)
to the real world. % --> better performance?.
Our approach outperforms model-free approaches in terms of grasping success rate
%(for grasping)
and does contrary to analytical approaches not require any human involvement (automatic configuration).
We demonstrate that our approach can be used for precise real-world robotic \mbox{pick-and-place} tasks, although being entirely trained on simulated data.

In future work, we plan to extend the approach to mixed bins
% and different gripper types (already part...)
and study how to reduce the time for data generation and training to allow a faster deployment of our solution. Furthermore, we want to study whether the generated data (grasping trials) can be used for model-free robotic grasping approaches in 6D.


%%%%%%%%%%%%%%%%%%%%%%%%%%%%%%%%%%%%%%%%%%%%%%%%%%%%%%%%%%%%%%%%%%%%%%%%%%%%%%%%

%\section*{ACKNOWLEDGMENT}
\section*{Acknowledgment}

% The preferred spelling of the word ÒacknowledgmentÓ in America is without an ÒeÓ after the ÒgÓ. Avoid the stilted expression, ÒOne of us (R. B. G.) thanks . . .Ó  Instead, try ÒR. B. G. thanksÓ. Put sponsor acknowledgments in the unnumbered footnote on the first page.

This work was partially supported by the
%Baden-W\"urttemberg Stiftung gGmbH (Deep Grasping -- Grant No. NEU016/1) and
% Bundesministerium für Bildung und Forschung (BMBF)
Federal Ministry of Education and Research (Deep Picking -- Grant No. 01IS20005C) and
the Ministry of Economic Affairs of the state Baden-W\"urttemberg (Center for Cognitive Robotics –- Grant No. 017-180004 and Center for Cyber Cognitive Intelligence (CCI) -- Grant No. 017-192996). We would like to thank our colleagues for helpful discussions
% and comments
and D. Unruh for the support with experiments.

% CCI/Arena2036 Quick Check mit TSE
% Deep Picking (zukünftig)
% KogRob (zukünftig)

% This work was partially supported by the Ministry of Economic Affairs of the state Baden-Württemberg (Zentrum für Kognitive Robotik – Grant No. 017-180004 and Zentrum für Cyber Cognitive Intelligence (CCI) – Grant No. 017-192996).

%%%%%%%%%%%%%%%%%%%%%%%%%%%%%%%%%%%%%%%%%%%%%%%%%%%%%%%%%%%%%%%%%%%%%%%%%%%%%%%%

\bibliographystyle{IEEEtran}
\bibliography{IEEEabrv,references}

\end{document}